%% file: main.tex
\documentclass[letterpaper,10pt,conference]{ieeeconf}  

\IEEEoverridecommandlockouts                              

\usepackage[noadjust]{cite}
\usepackage{xspace}
\input{math_commands.tex}
\usepackage{times}
\usepackage{epsfig}
\usepackage{graphicx}
\usepackage{amsmath}
\usepackage{amssymb}
\usepackage{booktabs}
\usepackage{multirow}
\let\labelindent\relax
\usepackage{enumitem}
\usepackage{array,makecell}
\usepackage[dvipsnames,table,xcdraw]{xcolor}
\definecolor{TableGreen}{RGB}{90, 175, 51}
\definecolor{mydarkblue}{RGB}{31,120,180}
\usepackage[pagebackref=true,breaklinks=true,colorlinks,bookmarks=false]{hyperref}
\hypersetup{
  citecolor=mydarkblue,
}

\usepackage{caption} 
\title{\LARGE \bf
Better Monocular 3D Detectors with LiDAR from the Past
}

\author{Yurong You$^{*\dag}$, Cheng Perng Phoo$^{*\dag}$, Carlos Andres Diaz-Ruiz$^{\ddag}$, Katie Z Luo$^\dag$, \\
Wei-Lun Chao$^\S$, Mark Campbell$^\ddag$, Bharath Hariharan$^\dag$, 
Kilian Q Weinberger$^\dag$%
\thanks{$*$ Equal contributions}%
\thanks{$\dag$ Computer Science Department, Cornell University 
        {\tt \{yy785, cp598, kzl6, bh497, kqw4\}@cornell.edu}}%
\thanks{$\ddag$ Mechanical and Aerospace Engineering Department, Cornell University 
        {\tt \{cad297, mc288\}@cornell.edu}}%
\thanks{$\S$ Department of Computer Science and Engineering, Ohio State University 
        {\tt chao.209@osu.edu}}%
}

\usepackage[capitalize]{cleveref}
\crefname{section}{Sec.}{Secs.}
\Crefname{section}{Section}{Sections}
\Crefname{table}{Table}{Tables}
\crefname{table}{Tab.}{Tabs.}

\newcommand{\loc}{\textit{p}\xspace}
\newcommand\mypara[1]{\vspace{.5mm}\noindent\textbf{#1}}

\newcommand{\lyft}{Lyft\xspace}
\newcommand{\ith}{Ithaca365\xspace}

\newcommand{\methodshort}{AsyncDepth\xspace}

\definecolor{iosgreen}{RGB}{83, 215, 105}
\definecolor{iosorange}{RGB}{253, 148, 38}
\definecolor{iosblue}{RGB}{20, 126, 251}

\makeatletter
\DeclareRobustCommand\onedot{\futurelet\@let@token\@onedot}
\def\@onedot{\ifx\@let@token.\else.\null\fi\xspace}

\def\eg{\emph{e.g}\onedot} 
\def\ie{\emph{i.e}\onedot} 
 
\def\etc{\emph{etc}\onedot}

\makeatother

\begin{document}

\maketitle
\thispagestyle{empty}
\pagestyle{empty}

\input{sections/0_abstract.tex}
\input{sections/1_introduction.tex}
\input{sections/2_related_work.tex}
\input{sections/3_methodology.tex}
\input{sections/4_experiments.tex}
\input{sections/5_discussion_conclussion.tex}

\vspace{-3px}

\section*{Acknowledgment}
\vspace{-3px}
This research is supported by grants from the US NSF (IIS-1724282, TRIPODS-1740822, IIS-2107077, OAC-2118240, OAC-2112606 and IIS-2107161), the ONR DOD (N00014-17-1-2175) and the Cornell Center for Materials Research with funding from the NSF MRSEC program (DMR-1719875). Katie Luo was supported in part by an Nvidia Graduate Fellowship.


\bibliographystyle{IEEEtran}
\bibliography{main}
\input{supp}

\end{document}

%% file: math_commands.tex
\usepackage{amsmath,amsfonts,bm}

\def\eqref#1{equation~\ref{#1}}

\def\1{\bm{1}}

\DeclareMathAlphabet{\mathsfit}{\encodingdefault}{\sfdefault}{m}{sl}
\SetMathAlphabet{\mathsfit}{bold}{\encodingdefault}{\sfdefault}{bx}{n}

\DeclareRobustCommand{\eg}{e.g.\@\xspace}
\DeclareRobustCommand{\ie}{i.e.\@\xspace}

%% file: sections/0_abstract.tex
\begin{abstract}
Accurate 3D object detection is crucial to autonomous driving. 
Though LiDAR-based detectors have achieved impressive performance, the high cost of LiDAR sensors precludes their widespread adoption in affordable vehicles. 
Camera-based detectors are cheaper alternatives but often suffer inferior performance compared to their LiDAR-based counterparts due to inherent depth ambiguities in images. 
In this work, we seek to improve monocular 3D detectors by leveraging unlabeled historical LiDAR data. Specifically, at inference time, we assume that the camera-based detectors have access to multiple unlabeled LiDAR scans from past traversals at locations of interest (potentially from other high-end vehicles equipped with LiDAR sensors).
Under this setup, we proposed a novel, simple, and end-to-end trainable framework, termed \methodshort, to effectively extract relevant features from asynchronous LiDAR traversals of the same location for monocular 3D detectors. We show consistent and significant performance gain (up to 9 AP) across multiple state-of-the-art models and datasets with a negligible additional latency of 9.66 ms and a small storage cost. Our code can be found at  \url{https://github.com/YurongYou/AsyncDepth}
\end{abstract}

%% file: sections/1_introduction.tex
\section{Introduction}
\label{sec:intro}
To drive safely, autonomous vehicles and driver assist systems must detect traffic participants and obstacles accurately. 
Current state-of-the-art prototypes rely on LiDAR sensors that provide accurate 3D information\cite{liu2022bevfusion}.
However, LiDAR sensors are expensive and their high cost precludes their mass adoption in consumer cars. 
Most commercially available driver assist systems instead rely on  
cheaper sensors --- (360$^{\circ}$-view) monocular cameras. Although more affordable, image-based 3D object detectors substantially underperform their LiDAR-based counterparts due to the inherent difficulty of inferring depth from images \cite{Qian_2020_CVPR}.

While it may be impractical and cost-prohibitive for \emph{every}  vehicle to be equipped with LiDAR sensors, \emph{some} (e.g. high-end luxury, police, etc) vehicles within a community may be outfitted with such sensors. 
In this setting, a few LiDAR-equipped vehicles collect data and share them (anonymously) with a large fleet of camera-only cars. If a camera-only car traverses a route for which past LiDAR data is available, it can fuse this data with its own sensor readings. A natural question to ask is: \emph{can we improve camera-based 3D object detectors using LiDAR data from the same location, but collected in the past?}

\begin{figure}[t]
    \centering
    \includegraphics[width=\linewidth]{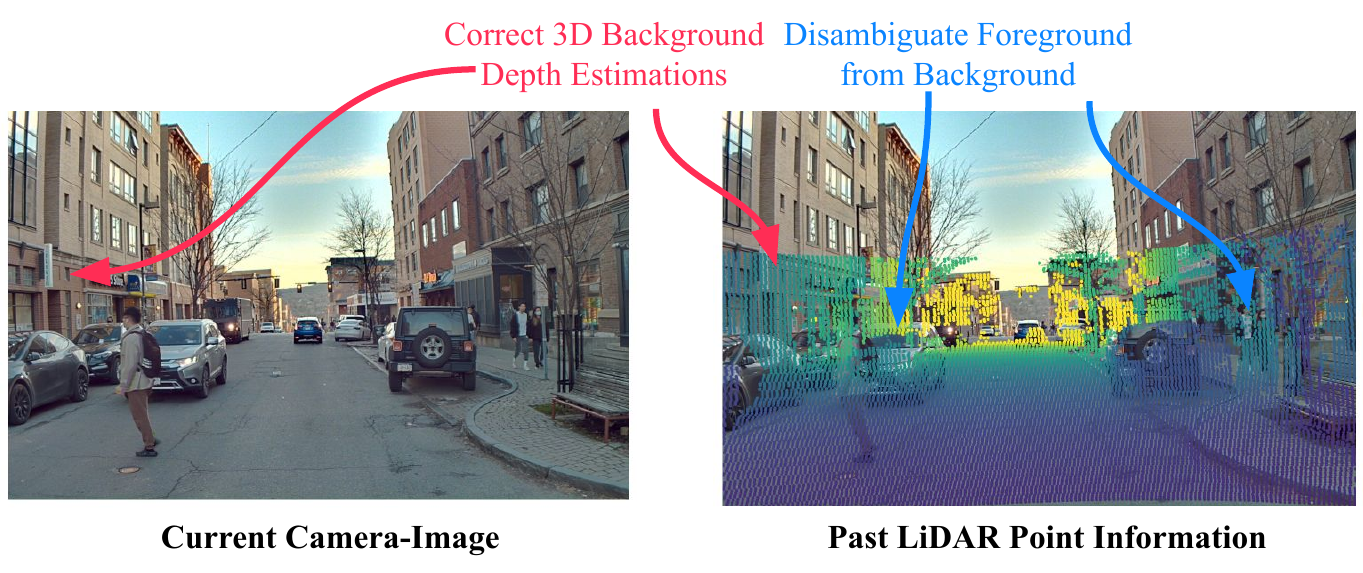}
    \caption{
    \textbf{Can past LiDAR traversals help monocular 3D object detection?} 
    Here we show a current image (left) and an asynchronous depth map rendered from a past LiDAR traversal (right).
    The asynchronous depth map provides accurate depth for background regions ({\color[HTML]{ff2d55}red} arrows) and 
    helps the monocular model disambiguate foreground objects in current scene ({\color[HTML]{0A84FF}blue} arrows).
    }
    \label{fig:teaser}
    \vspace{-25pt}
\end{figure}

Prior work has shown that combining 3D LiDAR point clouds with 2D images can improve 3D object detection
\cite{2022TransFusion,2022Distillation,You2020PseudoLiDARpp}, but these models crucially relies on a \textit{synchronized} LiDAR and camera sensors. In our case, however, the 3D data we have comes from a \emph{different car} passing through the scene presumably at a \emph{different time}. As such, vehicles and pedestrians will obviously have moved in the interim. Since these are the objects we want to detect in the current scene, these asynchronous offline LiDAR scans will not capture the shape and location of these objects.

However, even though the objects of interest may not be present in these past LiDAR scans, we argue that these scans still contain vital information for accurate 3D object detection. 
By aggregating data across multiple traversals, we can identify and remove transient  objects~\cite{barnes2018driven} and thus obtain accurate 3D data about the static background.
We posit that this 3D information about the background, collected from past traversals, can then be used to both \emph{detect} foreground objects in the current image as well as \emph{localize} them.
First, because foreground objects move and are therefore \emph{transient}, they will correspond to regions where the current image data is inconsistent with the previously collected depth ({\color[HTML]{0A84FF}blue} arrows in \autoref{fig:teaser}).
This can help the model detect ambiguous or partially occluded objects.
Second, in the areas where the previously collected depth is consistent with the current image, (\ie, the background), we get accurate depth for free ({\color[HTML]{ff2d55}red} arrows).
This accurate depth can be used by the image detector to localize foreground objects in 3D, \eg, by reasoning about where the pedestrian's feet meet the road.

Based on this insight, we propose a simple and effective approach for combining these asynchronous 3D data from past traversals with image-based detectors. We project each of the point clouds aggregated at each location into a depth map for each camera~\cite{Xu_2019_ICCV, Zhang_2018_CVPR}. 
From these depth maps, features are extracted, pooled across all past traversals, and combined with the image representation as the (intermediate) input to the monocular detector.
During training, the depth-map-based feature extractor is trained jointly with the object detector. During inference, the camera-only model can use the features extracted from past LiDAR scans to better detect and localize the objects.

We validate our approach, termed \methodshort, across two real-world self driving datasets, Lyft L5 Perception~\cite{lyftl5dataset} and Ithaca365~\cite{Diaz-Ruiz_2022_CVPR}, with two representative camera-based 3D object detection models~\cite{LSS2020,huang2021bevdet,liu2022bevfusion,FCOS3D2021,reading2021categorical}. 
Using our method, we observe a consistent improvement across both datasets, and up to 9.5 mAP over the baselines on far away ranges.
Our contributions are as follows:
\vspace{-3pt}
\begin{itemize}[itemsep=0pt]
    \item We study a novel yet highly practical scenario where \emph{asynchronous} historical LiDAR point cloud data is available to \emph{camera-only} perception systems.
    \item We show the practicality by proposing a simple and general approach to integrate asynchronous point cloud data into 3D monocular object detectors.
    \item We empirically demonstrate that our method yields consistent performance gains with low additional latency (9.66 ms) and a tiny storage cost across different datasets, detection ranges, object types, and detectors.
\end{itemize}
\vspace{-3pt}

%% file: sections/2_related_work.tex
\begin{figure*}[ht]
    \centering
    \includegraphics[width=.85\linewidth]{
    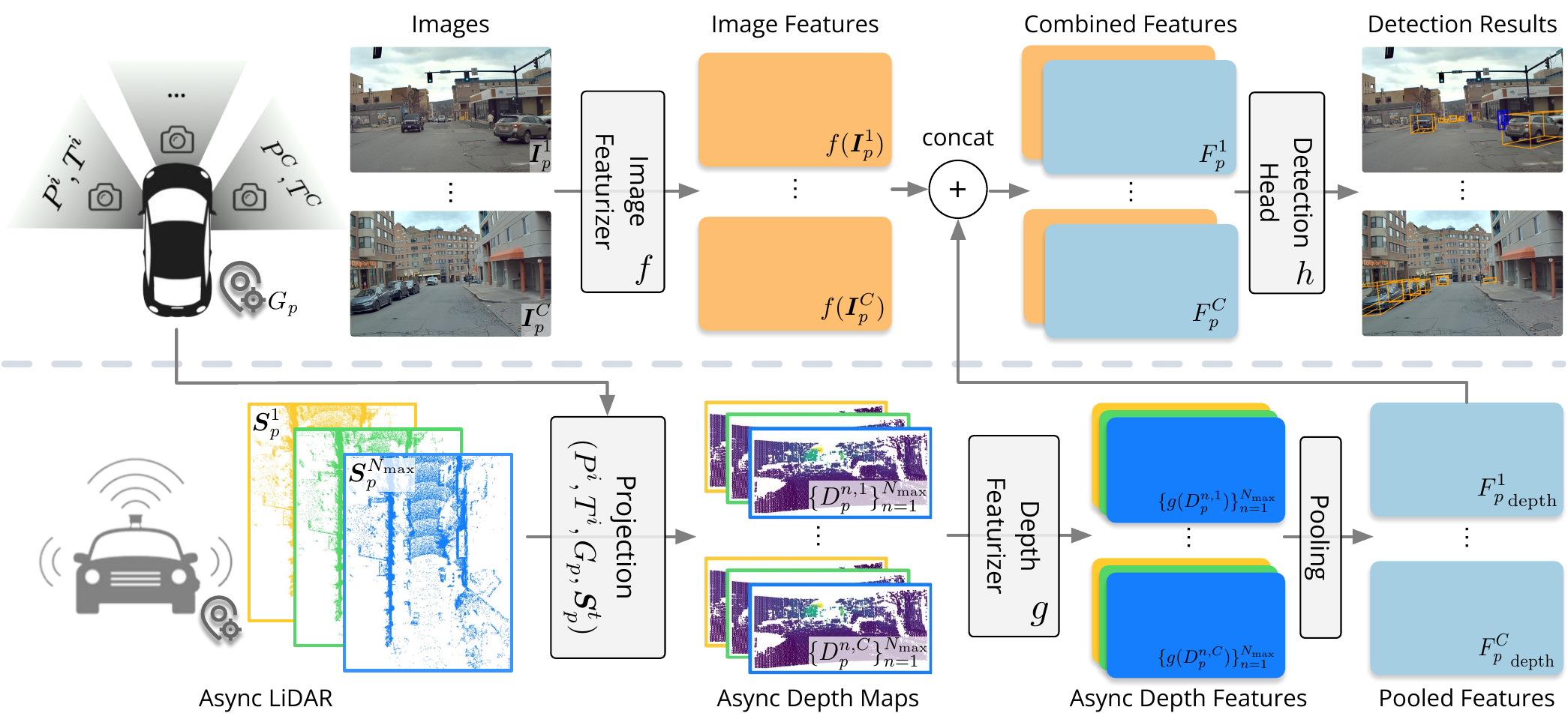
    }

    \vspace{-5pt}
    \caption{
    \textbf{Overview of \methodshort.} It consists of three parts: (top left) general ``featurize-then-detect'' pipeline for monocular 3D detection; 
    (bottom) extracting asynchronous depth features from past LiDAR traversals of the same location; 
    (top right) fusing the image features with \methodshort features. Please refer to \autoref{sec:method} symbol definitions.
    }
    \label{fig:pipeline}
    \vspace{-18pt}
\end{figure*}

\section{Related Work}
\label{sec:related}
\mypara{Perception for Autonomous Vehicles.}
Sensing the environment around a vehicle can be done via different input modalities, such as LiDAR or camera.  
LiDAR sensors are more expensive than cameras but are capable of capturing 3D geometry of the traffic scenes at high fidelity. Current state-of-the-art 3D object detectors therefore mostly rely on LiDAR sensors \cite{qi2017pointnet++,zhou2018voxelnet,shi2019pointrcnn,lang2019pointpillars,yang20203dssd}. Camera-based 3D object
detectors are a cheaper but also less accurate alternative due to depth ambiguities induced by perspective projections. The use of stereo-cameras~\cite{2019PseudoLidar,Li_2019_CVPR,You2020PseudoLiDARpp,Qian_2020_CVPR,PLUMENet2021} can close the gap, however the most common sensors in end user cars are monocular cameras \cite{mono3DChen2016,m3drpn2019,LSS2020,FCOS3D2021,li2022bevformer,wang2022probabilistic,Park_2021_ICCV,zhang2022beverse,xie2022m,liu2022petr,liu2022petrv2,wang2022detr3d,chen2022graph,reading2021categorical,2019PseudoLidar}, as they can be easily integrated within the car to capture a full 360 view around the vehicle. 
These monocular-based models can be roughly divided into two categories based on whether the detection is performed in 2D perspective view~\cite{mono3DChen2016,m3drpn2019,FCOS3D2021,wang2022probabilistic,Park_2021_ICCV} or in 3D~\cite{LSS2020,li2022bevformer,zhang2022beverse,xie2022m,liu2022petr,liu2022petrv2,wang2022detr3d,chen2022graph,reading2021categorical,2019PseudoLidar}.
In this work we show that both types of these models can be vastly improved through the use of offline LiDAR scans from past traversals. 

\mypara{Sensor Fusion in 3D Object Detection.}
LiDAR sensors yield accurate 3D geometry but suffer from sparse resolution; cameras, on other other hand, provide high resolution input but are inept in capturing 3D information. Given their complementary characteristics, multiple research efforts have explored fusing LiDAR and images for better 3D object detection \cite{han2017road,Vora2020PointPaintingSF,2022TransFusion,2022Distillation}. In contrast to our work, these approaches typically assume  \emph{synchronous} sensors and still require expensive LiDARs during inference. You et al.~\cite{You2020PseudoLiDARpp} address the cost issue by proposing to correct camera based perception through sparser (4-beams) and therefore cheaper LiDAR sensors. 
We explore an alternative setup, where the current scene is only captured by cheap cameras but \emph{asynchronous} LiDAR scans from the past are available. In principle, this is a much harder setup but it is of great practical value as it allows all cars to benefit from the LiDAR scans of a few (expensive) vehicles.  

\mypara{LiDAR Scans from Past Traversals. }
With accurate GPS/INS, LiDAR data from past traversals can be geo-located and aligned for easy retrieval. Recent work has started to explore the use of such data to aid perception and visual odometry~\cite{barnes2018driven}.  
 MODEST~\cite{you2022learning} leverages past traversals to discover dynamic objects without any annotations. Rote-DA~\cite{you2022unsupervised}  utilizes previous traversals to adapt pre-trained detectors to new target domains. You et al.~\cite{you2022hindsight} propose the use of past LiDAR point clouds to create feature descriptors for improving LiDAR-based 3D object detectors. 
 Nonetheless, these prior works demonstrate that past traversals are useful for various tasks in autonomous driving. We leverage this observation but in contrast, we are the first to show how to utilize past LiDAR to improve camera-only 3D object detection --- a common setting in practice. 

%% file: sections/3_methodology.tex
\section{\methodshort}

\mypara{Setup.}
We follow a typical test-time \emph{camera-only} sensor setup: 
the autonomous vehicle is equipped with synchronized sensors, including $C$ calibrated cameras and localization sensors (\eg, GPS/INS) but no LiDAR sensors.
The $C$ cameras are calibrated and have corresponding intrinsic and extrinsic matrices $\{(P^i, T^i)\}$.
These cameras capture $C$ images of the surrounding environment at a certain frequency.
When the vehicle drives through a location \loc, we denote the instantaneous images captured by the cameras as $\{\bm{I}_\loc^{i} \in \mathbb{R}^{H\times W\times 3}, i=1,\dots,C\}$. We also record the global 6-DoF localization as a rigid transformation $G_\loc$ that maps from the local to the global coordinate frame.
We do not assume that the fields of view of the $C$ cameras overlap, so work aims to develop a monocular 3D object detector that can identify objects of interest (dynamic traffic participants like cars, pedestrians, \etc) and infer their 3D positions, orientations, and sizes in the scene from these images. 

\mypara{Detector abstraction. } Current state-of-the-art monocular 3D object detectors~\cite{mono3DChen2016,m3drpn2019,LSS2020,FCOS3D2021,li2022bevformer} mostly follow a “featurize-then-detect" pipeline. Given images $\{\bm{I}_\loc^{i}\}$ and the corresponding camera parameters $\{(P^i, T^i)\}$, the detector first extracts image features $\{f(\bm{I}_\loc^{i})\}$ from each of the images via a (usually pre-trained) featurizer $f$;
a detector head $h$ then lifts these 2D feature maps to 3D object bounding boxes $\mathcal{B}_\loc = h(\{f(\bm{I}_\loc^{i}), P^i, T^i\})$ (here $f$ and $h$ contain learnable parameters). 
Such lifting from 2D to 3D is a notoriously \emph{ill-posed} problem for monocular detectors, since accurate \emph{depth} cannot be measured geometrically from the given 2D images~\cite{hartley2003multiple}. 
Current detectors get around this issue by learning a prior over depth~\cite{LSS2020,reading2021categorical,FCOS3D2021,li2022bevformer,wang2022detr3d}, but without any geometric or multi-view information, accurate depth estimation can be challenging.

\mypara{Overview of our approach.}
We propose a novel feature learning approach, termed \methodshort, to extract additional geometric information from past LiDAR scans that complements the image features $\{f(\bm{I}_\loc^{i})\}$ for 3D monocular object detection (\autoref{fig:pipeline}). 
We start by constructing asynchronous depth maps from the historical LiDAR point clouds using the localization and camera parameters. 
We then featurize these depth maps using a featurizer $g$ and aggregate them across traversals. 
The aggregated features are appended as additional channels to the image features for the detector head $h$. The whole pipeline is fully differentiable and can be learned end-to-end alongside almost all state-of-the-art monocular 3D object detectors.

\subsection{Past LiDAR Traversals}
\mypara{Past LiDAR traversals from other vehicles.} 
We assume offline data-sharing among vehicles equipped with \emph{different sensor modalities}. These vehicles drive about the same areas, collect \emph{unlabeled} sensor data, and share it with other vehicles when they are not operating.
Specifically in this work, we focus on one particular setting: vehicles with camera-only setups have access to past traversal data from vehicles with LiDAR sensors (this can be realized via community sharing or as a service provided by a vehicle manufacturer). 
These past LiDAR point clouds, though not capturing information about the instantaneous \emph{dynamic} objects in the current drive, can provide abundant 3D information about the \emph{static} environment.
It has been shown by \cite{you2022hindsight} that LiDAR-based detection models can be enhanced by these past traversals.

Different from \cite{you2022hindsight}, we assume a camera-only setup for the operating vehicle. 
We hypothesize that the environment information within these past LiDAR traversals can 
greatly help monocular models detect objects.
We validate this hypothesis with a generally applicable and simple framework for monocular 3D detectors. 

\mypara{LiDAR Densification. } We follow \cite{you2022hindsight} and maintain a maximum of $N_{\max} \geq 1$ LiDAR traversals for the driving locations (bottom half of \autoref{fig:pipeline}). Each traversal $n$ is a sequence of point clouds $\{\bm{Q}_r^n \in \mathbb{R}^{k\times 3}\}$, where $r$ index the frame and $k$ is the number of points, obtained from the past traversals of other vehicles. We transform each point cloud into the same global coordinate system via the associated 6-DoF localization. We combine point clouds along the road to obtain a densified point cloud $\bm{S}^n_\loc = \bigcup_{r \in R}\bm{Q}_r^n$, where $R$ is a subset of frames sampled every $s$ meters along the road near location \loc. 
Of course, these point clouds contain both dynamic objects in the past and static background. But as pointed out by \cite{barnes2018driven,you2022learning}, with multiple traversals of the same scene, simple statistics (\ie ephemerality/persistency point score) can already help to disambiguate dynamic/static components. Rather than constructing hand-crafted statistics from these point clouds, we propose to learn a feature extractor that can extract relevant information from them. 
\subsection{Asynchronous Depth Feature from Past LiDAR} 
As discussed previously, inferring depth information from images is an ill-posed problem. 
However, the densified point clouds $\{\bm{S}^n_{\loc}\}_{n=1}^{N_{\max}}$ at location \loc can provide strong cues for estimating the object position.
We project this point cloud into each camera's image to yield a corresponding depth map.
Concretely, we use the current 6-DoF localization $G_\loc$ and the $i$-th camera's parameters $(P^i, T^i)$ to perform this projection.
For every 3D point in the densified point cloud $\bm{S}^n_{\loc}$ (represented in homogeneous coordinates as $\bm{q}_j \equiv [x_j, y_j, z_j, 1]^\top$), we project it to the local camera coordinate,
\begin{equation*}
    \hat{\bm{q}}_j =  [\hat{x}_j, \hat{y}_j, \hat{z}_j, 1]^\top \equiv T^i G_\loc^{-1} \bm{q}_j.
\end{equation*}
We then project each of these points $\hat{\bm{q}}_j$ onto the image plane of the $i$-th camera by perspective projection:
\begin{align*}
    [\hat{u}_j^i \hat{z}_j, \hat{v}_j^i \hat{z}_j, \hat{z}_j]^\top & \equiv P^i \hat{\bm{q}}_j \\
    u_j^i, v_j^i &= \lfloor \hat{u}_j^i\rfloor, \lfloor \hat{v}_j^i\rfloor  
\end{align*}
where $(u_j^i, v_j^i)$ are the corresponding pixel indices on the $i$-th image. 
Projecting all points in $\bm{S}^n_{\loc}$ into the image plane of camera $i$ and filling the corresponding depth value renders a depth map $D_\loc^{n, i} \in \mathbb{R}^{H\times W}$ (see the right image of \autoref{fig:teaser})  
\begin{equation*}
\textstyle
    D_\loc^{n, i}[u,v] = \max\left\{\max_{(u_j^i, v_j^i) = (u, v)}\hat{z}_j, -1 \right\},
\end{equation*}
where we take the maximum depth when multiple points are projected into a same pixel and fill the empty pixel with -1. This implicitly favors background depth since foreground depth is usually closer. As shown in \autoref{fig:teaser}, such a depth map provides very rich depth prior for the image. 
Intuitively, once the model detects an object in 2D, figuring out its depth is a much easier task with the surrounding \emph{background} depth.

\subsection{Feature Learning and Detection}

The previous depth maps, $\{D_\loc^{n, i}\}$, can be noisy since they were captured under different conditions and contain different sets of prior foreground objects.
To extract relevant information, we feed them through a 2D backbone $g$, which is designed to yield depth feature maps $g(D_\loc^{n, i}) \in \mathbb{R}^{H'\times W' \times d_\text{depth}}$ with the same size as that of the image features $f(\bm{I}^i_p)$. 
To aggregate the depth features from different traversals, we apply an order invariant pooling function to pool the feature maps along the traversal dimension:
\begin{equation*}
\textstyle
    {F_\loc^i}_\text{depth}[u,v] = \text{pool}(g(D_\loc^{n, i})[u,v], n=1,\dots,N_{\max} ).
\end{equation*}
We use mean pooling in our implementation by default. 
The pooled feature maps from the past LiDAR traversals are concatenated with the corresponding image features along the feature dimension as the new 2D features ${F_\loc^i} = \text{concat}({F_\loc^i}_\text{depth}, f(\bm{I}^i_p))$. We then apply the same detector head $h$ (with slight change to input feature size) to obtain the bounding box predictions $\mathcal{B}_\loc = h(\{{F_\loc^i}, P^i, T^i\})$.
As a result, the information from past LiDAR traversals can be incorporated into the existing camera-only object detection models with minimal changes in model architecture.

\mypara{Training and Inference.} 
We train the whole model, including (i) the depth backbone $g$ that takes LiDAR scans of \emph{past} traversals as input, (ii) the image featurizer $f$ that takes images at the \emph{current} time as input, and (iii) the detector head $h$, end-to-end  with loss signals from annotated 3D labels of the \emph{current} scene. We keep the loss designs of baseline camera-based detection models intact.

During inference, we assume that the model has access to the past LiDARs traversals and generates the depth map online.
The depth map backbone can run in parallel with the image featurizer to reduce latency.

\label{sec:method}

%% file: sections/4_experiments.tex
\section{Experiments}
\label{sec:experiments}

\input{tables/lyft_main}
\input{tables/ithaca_main}
\mypara{Datasets.}
We validate our approach on two large-scale datasets: Lyft L5 Perception Dataset~\cite{lyftl5dataset} and \ith~\cite{Diaz-Ruiz_2022_CVPR} Dataset. 
To the best of our knowledge,
these are the only two publicly available autonomous driving datasets that have both bounding box annotations and multiple traversals with accurate localization (Note that nuScenes~\cite{caesar2020nuscenes}  contains some scenes with multiple traversals but the localization in $z$-axis is not accurate ~\cite{nusc_inaccurate_localization}). 
The \lyft dataset is collected in Palo Alto (California) and the \ith dataset is collected in Ithaca (New York).
Both datasets provide camera images (6 ring-camera images in \lyft and 2 frontal-view images in \ith) and 3D LiDAR scans (40-beam in \lyft and 128-beam in \ith).
We thus perform 360-degree detection on \lyft and frontal-view only detection on \ith.
The detection range is set to maximum $50$m to the ego vehicle, following the setup of most camera-only detection models developed on nuScenes dataset~\cite{caesar2020nuscenes}.
For the \lyft dataset, to ensure fair assessment of generalizability, we re-split the dataset so that the training set and test set are geographically disjoint; we also discard locations with less than 2 traversals in the training set. This results in a train/test split of 10,499/3,412 examples.
For the \ith dataset, we follow the default split of the dataset, which results in 4,445/1,644 train/test examples covering the same route but with different collection times.
Adhering to our setup, synchronized LiDAR point clouds are \textbf{not used} during testing.

\mypara{Localization.} With current localization technology, we can achieve high localization accuracy (\eg, 1-2 cm level accuracy with RTK). We assume good localization in asynchronous LiDAR traversals and the camera-only systems and study the effect of the localization error in the supplementary. 

\mypara{Evaluation metric.} We adopt similar metrics from the nuScenes dataset~\cite{caesar2020nuscenes} to evaluate the detection performance. We evaluate detection performance within $50$m of the ego vehicle. 
The mean average precision (mAP) is the mean of average precisions (AP) of different classes under $\{0.5, 1, 2, 4\}$m thresholds that determine the match between detection and ground truth. 
Because \lyft and \ith datasets do not provide the objects' velocities and attributes ground-truths, we only compute 3 types of true positive metrics (TP metrics), including ATE, ASE and AOE for measuring translation, scale and orientation errors. These TP metrics are computed under a match distance threshold of $2$m. 
Additionally, we also compute a distance based breakdown ($0$-$30$m, $30$-$50$m) for these metrics. We evaluate 5 foreground objects (car, truck, bus, pedestrian and bicycle) on \lyft and 2 objects (car, pedestrian) on \ith. Similar to NDS (nuSccenes detection score), we calculate the overall detection score (DS) for these two datasets as $\text{DS}=\frac{1}{6}[3\cdot\text{mAP} + \sum_{\text{mTP} \in \mathbb{TP}} (1 - \min(1, \text{mTP}))]$. 
To showcase the most significant improvements from \methodshort, we mainly present mAP evaluation results on the main paper and include the rest in the supplementary due to space limitations.

\mypara{Detection models.} 
We experiment with two representative, high-performing monocular 3D object detection models: FCOS3D \cite{FCOS3D2021} and Lift-Splat~\cite{LSS2020,huang2021bevdet,reading2021categorical,liu2022bevfusion}. 
FCOS3D extends 2D object detection~\cite{tian2019fcos} to 3D by detecting objects in perspective views and regressing additional 3D targets for each of the detected objects. 
The Lift-Splat style model first constructs a Bird's Eye View (BEV) representation of the scene and then applies a detection head. This BEV representation is constructed by predicting the depth distribution for each pixel on the image feature map and ``splatting'' the corresponding weighted 2D features into BEV space via camera parameters. 
Thanks to their strong performance and clean design, these two types of monocular 3D object detection models have been well-received by the community~\cite{wang2022probabilistic,Park_2021_ICCV,zhang2022beverse,xie2022m,liu2022petr,liu2022petrv2}.

\mypara{Implementation details.} 
We strive for a clean and simple implementation to show the generalizability of the proposed approach: 
we adopt an official implementation~\cite{mmdet3d2020} of these two models with minimal changes only for supporting the \lyft and \ith datasets, and use the \emph{exact same} hyper-parameters on both datasets without bells and whistles. 
For the Lift-Splat / BEVDet model, we adopt the efficient camera-to-BEV transformation implementation in \cite{liu2022bevfusion} and a detection head similar to \cite{yin2021center}. We use a Swin-T~\cite{liu2021swin} pre-trained on nuImages~\cite{caesar2020nuscenes} as the image backbone, and supervise depth prediction by the smooth-L1 loss against ground-truth depth during training.
For the FCOS3D model, we follow the original paper and use the official training schedule. We use a pretrained ResNet101~\cite{deng2009imagenet,he2016deep} with deformable convolutions~\cite{dai2017deformable} for image feature extraction.
We use ResNet18 \cite{he2016deep} to extract features from the depth maps for the different 3D detectors. 
We deploy a feature pyramid network \cite{lin2017fpn} to extract multi-scale features if the 3D detector of interest is also using multi-scale features.
For both datasets, we use a maximum 5 other traversals at the reference location to obtain the depth maps. For the \lyft dataset, for each past traversal we use point clouds closest to $\{0, -20, 20\}$m to the ego vehicle along the road since we are performing 360-degree detection; for the \ith dataset, we use $\{0, 10, 20\}$m for frontal-view detection. 

\subsection{Monocular 3D Detections with \methodshort}

We show the performance of various detectors with and without \methodshort on \lyft and \ith in \autoref{tab:lyft_map} and \autoref{tab:ithaca_map} respectively. Overall, we observe that using LiDAR scans from past traversals can significantly improve the performance of monocular 3D object detectors. On \lyft, we observe an improvement of 1.8 mAP averaged across different detectors and different classes at various evaluation thresholds. On \ith, we observe an even more pronounced improvement over the baselines (the \methodshort variants outperform the baseline by an average of 3.9 mAP). 

To understand where the performance gains are from, we look at the performance of the detectors on various classes. On \lyft, we observe that the biggest improvements come from the detection of bus and bicycles (with an improvement of 6 points and 1.9 respectively); on \ith, the performance gain of \methodshort largely reflects on car detection, with a remarkable 5.8 improvement in performance. 

In addition, we also look at the performance of the detectors at various ranges. The performance improvement from using \methodshort is most pronounced in the challenging far-range object detection (30-50m). On \lyft, we even observe a stark improvement of 9.5 mAP in far-range bus detection over the Lift-Splat baseline; on \ith, we observe an average improvement 6.9 over the baselines across the two detectors on car detection. This suggests that all the detectors benefit from the depth information encoded in the features, particularly in the far ranges where it is especially difficult to infer depth from images. 

\subsection{Ablation study}
\noindent All ablations are on the Lift-Splat detector on the \ith.

\mypara{Effect of using synchronous depth maps.}
\input{tables/gt_conditioning.tex}
Our approach involves learning complementary features from the asynchronous depth maps constructed from historical traversals. Discerning readers might question how our method would perform if we instead used synchronous depth maps for learning the feature extractors. 
In \autoref{tab:gt_conditioning_map}, we observe that learning features from synchronous depth map (+ SyncDepth) indeed outperforms the baseline by a significant margin, thus validating the claim that accurate depth is crucial to 3D object detection. 
Though leveraging offline depth maps (+\methodshort) is worse than using synchronous depth maps, synchronous depth maps requires real-time LiDAR sensing which is expensive to obtain. 
Offline depth maps are a cheaper alternative that can significantly boost the performance of detectors, greatly improving the sensing ability of camera-only autonomous vehicles.

\mypara{Effects of feature extractors.}
\input{tables/different_backbones.tex}
One key aspect of our approach is to deploy an image featurizer to featurize the asynchronous depth-maps before aggregating various information from different traversals. 
To validate such a design choice, we consider a non-learning baseline (Down. + Avg.) in which we first downsample each offline depth map to appropriate sizes using bilinear interpolation and average them to form a single channel feature that can be appended to the extracted image feature maps. 
We present the result in \autoref{tab: different_backbone_map}. Naively averaging the asynchronous depth-maps does bring forth some improvements over the baseline but it is far from using a learnable feature extractor. In addition, we also investigated the difference between using a pre-trained ImageNet ResNet18 and a randomly initialized ResNet18. Although the backbone has been pretrained on ImageNet, consisting of natural images, we observed improvements brought by this initialization, especially on car detections. This validates previous results in the literature \cite{
kolesnikov2020biT, 
zhai2019vtab, 
guo2020CDFSL, 
phoo2020startup, 
dosovitskiy2020ViT}.

\mypara{Different number of historical traversals.}
\input{tables/different_num_traversals.tex}
Throughout the text, we assume the max number of past traversals for each scene $N_{\max} \leq 5$. However, due to privacy concerns or hardware failures, we might have access to less than 5 during inference. To investigate the robustness of \methodshort, we conduct inference with various number of upper bound for $N_{\max}$ in \autoref{tab: different_num_traversals}. With just 2 traversals for each scene, \methodshort can outperform the baseline ($N_{\max}=0$) by a large margin (3.3 AP for car and 1.4 AP for pedestrian).  

\mypara{Data storage and latency.} 
We provide an analysis of the additional computational and storage overhead introduced by our method. 
On average, the AsyncDepth part yields an extremely low \emph{9.66 ms} latency (whole model: 70.78 ms, image featurizer: 23.94 ms).
This is due to the relatively small network (ResNet-18) in AsyncDepth. 
The latency can be further decreased since the depth featurizer can run in parallel with the image featurizer. 
For data storage and transmissions, the LiDAR points for 5 past traversals of a single scene take about 17.16 MB. For context, the average American commute about 15 miles to work on average~\cite{average_american_commute}. For typical usage, our method needs 13.49 GB to store 5 past traversals. The cost to store this amount of data is low --- with current technology, it costs about \$0.01/GB for hard drives --- and it can be further reduced with compression.

\mypara{Supplementary.} Please refer to the supplementary for more ablation studies and qualitative visualization.

%% file: tables/lyft_main.tex
\begin{table*}[!t]
    \tabcolsep 3pt
    \centering
    \resizebox{\textwidth}{!}{%
    \begin{tabular}{@{}ccccccccccccccccc@{}}
        \toprule 
        \multicolumn{1}{c}{\multirow{2}{*}{Method}} & \multicolumn{1}{c}{\multirow{2}{*}{mAP}} &  \multicolumn{3}{c}{Car} & \multicolumn{3}{c}{Truck} & \multicolumn{3}{c}{Bus} & \multicolumn{3}{c}{Bicycle} &  \multicolumn{3}{c}{Pedestrian} \\
        \cmidrule(lr){3-5} \cmidrule(lr){6-8} \cmidrule(lr){9-11} \cmidrule(l){12-14} \cmidrule(l){15-17}
        & & 0-30 & 30-50 & 0-50 & 0-30 & 30-50 & 0-50 & 0-30 & 30-50 & 0-50 & 0-30 & 30-50 & 0-50 & 0-30 & 30-50 & 0-50 \\
        \midrule
        FCOS3D~\cite{FCOS3D2021} & 14.6 & 47.4 & 23.6 & 37.9 & 4.5 & 3.1 & 4.2 & 5.1 & 3.9 & 5.1 & 20.3 & 1.3 & 10.1 & 25.7 & 3.6 & 15.7\\
        + \methodshort & 16.0 & 48.3 & 24.5 & 38.8 & 7.2 & 4.1 & 6.0 & 9.2 & 7.7 & 8.9 & 24.4 & 1.5 & 10.8 & 26.3 & 1.7 & 15.8 \\
        \cmidrule{2-17}
        $\Delta$ AP & \textcolor{TableGreen}{+1.4 }& \textcolor{TableGreen}{+0.9 }& \textcolor{TableGreen}{+0.9 }& \textcolor{TableGreen}{+0.9 }& \textcolor{TableGreen}{+2.7 }& \textcolor{TableGreen}{+1.0 }& \textcolor{TableGreen}{+1.8 }& \textcolor{TableGreen}{+4.1 }& \textcolor{TableGreen}{+3.8 }& \textcolor{TableGreen}{+3.8 }& \textcolor{TableGreen}{+4.1 }& \textcolor{TableGreen}{+0.2 }& \textcolor{TableGreen}{+0.7 }& \textcolor{TableGreen}{+0.6} & \textcolor{red}{-1.9} & \textcolor{TableGreen}{+0.1} \\
        \midrule
        Lift-Splat~\cite{LSS2020} & 23.3 & 65.2 & 25.6 &	50.7 &	11.9 &	5.6 &	9.9 &	18.7 &	13.3 &	15.7 &	31.3 &	0.3 & 13.7	& 35.7 & 4.2	& 21.3 \\
        + \methodshort & 25.4 & 66.7 & 27.0 & 52.2 & 14.5 & 6.9 & 11.0 & 22.4	& 22.8 &24.0& 	34.3& 	0.4& 	15.9& 	35.9& 	8.4& 	23.8 \\
         \cmidrule{2-17}
        $\Delta$ AP & \textcolor{TableGreen}{+2.1} & \textcolor{TableGreen}{+1.5}&	\textcolor{TableGreen}{+1.4}	&\textcolor{TableGreen}{+1.5}&	\textcolor{TableGreen}{+2.6}&	\textcolor{TableGreen}{+1.4}&	\textcolor{TableGreen}{+1.1}&	\textcolor{TableGreen}{+3.7}&	\textcolor{TableGreen}{+9.5}&	\textcolor{TableGreen}{+8.3}&	\textcolor{TableGreen}{+3.1}&	\textcolor{TableGreen}{+0.1}&	\textcolor{TableGreen}{+2.3}&	\textcolor{TableGreen}{+0.3}&	\textcolor{TableGreen}{+4.2}&	\textcolor{TableGreen}{+2.5} \\
        \bottomrule
    \end{tabular}
    }
    \vspace{-5pt}
    \caption{\textbf{Mean Average Precision (mAP) of two types of detectors across different ranges and object class types on the \lyft dataset.} 
    We evaluate two types of monocular 3D object detection models (FCOS3D~\cite{FCOS3D2021} and Lift-Splat~\cite{LSS2020,liu2022bevfusion}
    We show the mAP metric and its breakdown across different ranges (in meters) and class objects. 
    ``$\Delta$ AP'' indicates the gain.
    We observe \methodshort improves the reference detectors in all but one case.
    Other corresponding metrics (ATE, ASE, AOE and DS) are included in the supplementary material where we observe a similar trend.
    }
    \label{tab:lyft_map}
    \vspace{-15pt}
\end{table*}

%% file: tables/ithaca_main.tex
\begin{table}[!t]
    \centering
    \tabcolsep 3pt
    \resizebox{.47\textwidth}{!}{%
    \begin{tabular}{@{}cccccccc@{}}
        \toprule 
        \multicolumn{1}{c}{\multirow{2}{*}{Method}} & \multicolumn{1}{c}{\multirow{2}{*}{mAP}} & \multicolumn{3}{c}{Car} &  \multicolumn{3}{c}{Pedestrian} \\
        \cmidrule(lr){3-5} \cmidrule(lr){6-8}
        & & 0-30 & 30-50 & 0-50 & 0-30 & 30-50 & 0-50  \\
        \midrule
        FCOS3D~\cite{FCOS3D2021} & 25.0 & 46.3 & 23.8 & 36.2 & 18.3 & 7.8 & 13.8  \\
        + \methodshort & 29.2 & 51.7 & 29.6 & 42.2 & 20.0 & 10.3 & 16.2 \\
        \cmidrule{2-8}
        $\Delta$ AP & \textcolor{TableGreen}{+4.2} & \textcolor{TableGreen}{+5.4} & \textcolor{TableGreen}{+5.8} & \textcolor{TableGreen}{+6.0} & \textcolor{TableGreen}{+1.7} & \textcolor{TableGreen}{+2.5} & \textcolor{TableGreen}{+2.4}  \\
        \midrule
        Lift-Splat~\cite{LSS2020} & 39.4 & 66.6 & 30.2 & 52.8 & 37.1 & 13.7 & 26.0 \\
        + \methodshort & 42.9 & 70.2 & 38.2 & 58.3 & 37.6 & 16.6 & 27.5 \\
        \cmidrule{2-8}
        $\Delta$ AP & \textcolor{TableGreen}{+3.5} & \textcolor{TableGreen}{+3.6} & \textcolor{TableGreen}{+8.0} & \textcolor{TableGreen}{+5.5} & \textcolor{TableGreen}{+0.5} & \textcolor{TableGreen}{+2.9} & \textcolor{TableGreen}{+1.5}  \\
        \bottomrule
    \end{tabular}
    }
    \vspace{-5pt}
    \caption{\textbf{Mean Average Precision (mAP) of two types of detectors across different ranges and object types on the \ith dataset.} Please refer to \autoref{tab:lyft_map} for naming.
    \methodshort improves reference models in all cases.
    }
    \label{tab:ithaca_map}
    \vspace{-18pt}
\end{table}

%% file: tables/gt_conditioning.tex
\begin{table}[!t]
    \small
    \centering
    \tabcolsep 4pt
    \resizebox{.47\textwidth}{!}{%
    \begin{tabular}{cccccccc}
        \toprule 
        \multicolumn{1}{c}{\multirow{2}{*}{\thead{Lift-Splat\\ Variants}}} & \multicolumn{1}{c}{\multirow{2}{*}{mAP}} & \multicolumn{3}{c}{Car} &  \multicolumn{3}{c}{Pedestrian} \\
        \cmidrule(lr){3-5} \cmidrule(lr){6-8}
        & & 0-30 & 30-50 & 0-50 & 0-30 & 30-50 & 0-50  \\
        \midrule
        baseline       & 39.4 & 66.6 & 30.2 & 52.8 & 37.1 & 13.7 & 26.0 \\
        + \methodshort & 42.9 & 70.2 & 38.2 & 58.3 & 37.6 & 16.6 & 27.5 \\
        \textcolor{Gray}{+ SyncDepth} & \textcolor{Gray}{51.1} & \textcolor{Gray}{72.3} & \textcolor{Gray}{44.9} & \textcolor{Gray}{62.3} & \textcolor{Gray}{48.0} & \textcolor{Gray}{29.9} & \textcolor{Gray}{39.9} \\
        \bottomrule
    \end{tabular}
    }
    \vspace{-5pt}
    \caption{\textbf{Mean average precision for Lift-Splat
    model with asynchronous/synchronous depth map on \ith dataset.} 
    ``+ \methodshort'' stands for the proposed method using depth maps from asynchronous LiDAR. ``+ SyncDepth'' stands for an \emph{oracle} scenario where we replace asynchronous depth maps with synchronized depth.
    }
    \label{tab:gt_conditioning_map}
    \vspace{-10pt}
\end{table}

%% file: tables/different_backbones.tex
\begin{table}[!t]
    \small
    \centering
    \resizebox{.47\textwidth}{!}{%
    \tabcolsep 4pt
    \begin{tabular}{cccccccc}
        \toprule 
        \multicolumn{1}{c}{\multirow{2}{*}{\thead{Depth \\ Featurizer}}} & \multicolumn{1}{c}{\multirow{2}{*}{mAP}} & \multicolumn{3}{c}{Car} &  \multicolumn{3}{c}{Pedestrian} \\
        \cmidrule(lr){3-5} \cmidrule(lr){6-8}
        & & 0-30 & 30-50 & 0-50 & 0-30 & 30-50 & 0-50  \\
        \midrule
        N/A       & 39.4 & 66.6 & 30.2 & 52.8 & 37.1 & 13.7 & 26.0 \\\midrule
        Down.~+~Avg.  & 39.8 & 66.7 & 31.5 & 53.3 & \textbf{37.7} & 14.5 & 26.5 \\
        Random Init.  & 41.6 & 69.1 & 37.0 & 57.0 & 35.1 & 16.0 & 26.2 \\
        ImageNet Init. & \textbf{42.9} & \textbf{70.2} & \textbf{38.2} & \textbf{58.3} & 37.6 & \textbf{16.6} & \textbf{27.5} \\
        \bottomrule
    \end{tabular}
    }
    \vspace{-5pt}
    \caption{\textbf{Mean average precision of our methods with different depth featurizers.} 
    ``Down.~+~Avg.'' stands for directly using the downsampled projected asynchronous depth maps and averaging them across traversals; ``Random Init.'' and ``ImageNet Init.'' initialize the same featurizer randomly and from ImageNet pre-trained weights, respectively.}
    \label{tab: different_backbone_map}
    \vspace{-18pt}
\end{table}

%% file: tables/different_num_traversals.tex
\begin{table}[!t]
    \small
    \centering
    \resizebox{.47\textwidth}{!}{%
    \tabcolsep 4pt
    \begin{tabular}{cccccccc}
        \toprule 
        \multirow{2}{*}{\# Traversals} & \multicolumn{1}{c}{\multirow{2}{*}{mAP}} & \multicolumn{3}{c}{Car} &  \multicolumn{3}{c}{Pedestrian} \\
        \cmidrule(lr){3-5} \cmidrule(lr){6-8}
          & & 0-30 & 30-50 & 0-50 & 0-30 & 30-50 & 0-50  \\
        \midrule
        $N_{\max}=0$ & 39.4 & 66.6 & 30.2 & 52.8 & 37.1 & 13.7 & 26.0 \\
        \midrule
        $N_{\max} = 1$ & 40.2& 66.4 & 34.8 & 54.2 & 35.5 & 16.2 & 26.3 \\
        $N_{\max} \leq 2$  & 41.8 & 68.2 & 36.0 & 56.1 & 36.6 & 17.0 & 27.4 \\
        $N_{\max} \leq 5$  & 42.9 & 70.2 & 38.2 & 58.3 & 37.6 & 16.6 & 27.5\\
        \bottomrule
    \end{tabular}
    }
    \vspace{-5pt}
    \caption{\textbf{Mean Average Precision of using \methodshort with various number of past traversals during inference.} $N_{\max}=0$ corresponds to vanilla Lift-Splat baseline model without using past LiDAR traversals. $N_{\max}\leq m$ stands for only using $\leq m$ past traversals during testing.}
    \label{tab: different_num_traversals}
    \vspace{-18pt}
\end{table}

%% file: sections/5_discussion_conclussion.tex
\section{Conclusion}
We explore using asynchronous LiDAR scans from past traversals to improve monocular 3D detectors for autonomous vehicles. 
Though not containing information about the location/shape of the target objects in the current scene, we show that these LiDAR scans still contain vital information that can aid 3D object detection. 
Specifically, we extract offline depth maps from the past traversals and use these depth maps to learn features that aid monocular 3D object detectors. 
Our approach is simple, lightweight, and compatible with practically all state-of-the-art monocular detectors. We show consistent enhancement of multiple detectors on multiple datasets, opening up new possibilities in improving monocular 3D detection using past traversals.

%% file: supp.tex
\setcounter{equation}{0}
\setcounter{figure}{0}
\setcounter{table}{0}
\setcounter{page}{1}
\setcounter{section}{0}
\renewcommand{\thesection}{S\arabic{section}}
\renewcommand{\thetable}{S\arabic{table}}
\renewcommand{\thefigure}{S\arabic{figure}}

\begin{center}
  \textbf{\Large Supplementary Material}
\end{center}

\section{Implementation Details}
\mypara{Lift-Splat.} We adopt the implementation in BEVFusion\footnote{\url{https://github.com/mit-han-lab/bevfusion/blob/main/configs/nuscenes/det/centerhead/lssfpn/camera/256x704/swint/default.yaml}} for Lift-Splat model. We default train the whole model with 20 epochs by AdamW~\cite{loshchilov2018decoupled} with 4 GPUs and batch size 2 per GPU. 
We use a cyclic learning rate schedule which starts from $1\times 10^{-4}$, climbs to $1 \times 10^{-3}$ and anneals to $0$ in the training process (we also do a learning rate linear warm up from 0 for the first 500 steps to stabilize training). For image backbone, we use $1/10$ of the default learning rate to fine-tune. 
We downscale and crop the images to $384 \times 800$ for \lyft and $384 \times 896$ for \ith.
Since we only focus on image-only detection within $50$m around the ego vehicle, we limit the depth prediction range to $(0, 60)$ and every $0.5$m. During training, we supervise the depth prediction by the smooth-L1 loss between ground-truth / synchronous depth and the weighted sum of the predicted depth bins. All hyper-parameters are kept the same between the experiments on the two datasets.

\mypara{FCOS3D.} We adopt the implementation in MMDetection3D\footnote{\url{https://github.com/open-mmlab/mmdetection3d/blob/master/configs/fcos3d/README.md}} for FCOS3D models. We follow the training schedule in the original paper~\cite{FCOS3D2021}, where we first train the whole model with a weight of $0.2$ for depth regression and then fine-tune it with a weight of $1$. For initial training, we train 20 epochs by SGD with 4 GPUs and batch size 4 per GPU. We adopt a step learning rate, where the initial learning rate is $2\times 10^{-3}$ and down-scaled by $0.1$ on epoch 10 and 15. For fine-tuning, we change the initial learning rate to $1 \times 10^{-3}$ and also train 20 epochs. All hyper-parameters are kept the same between the experiments on the two datasets.

\section{Qualitative Analysis: Detection Results}
\begin{figure*}[t]
    \centering
    \includegraphics[width=\textwidth]{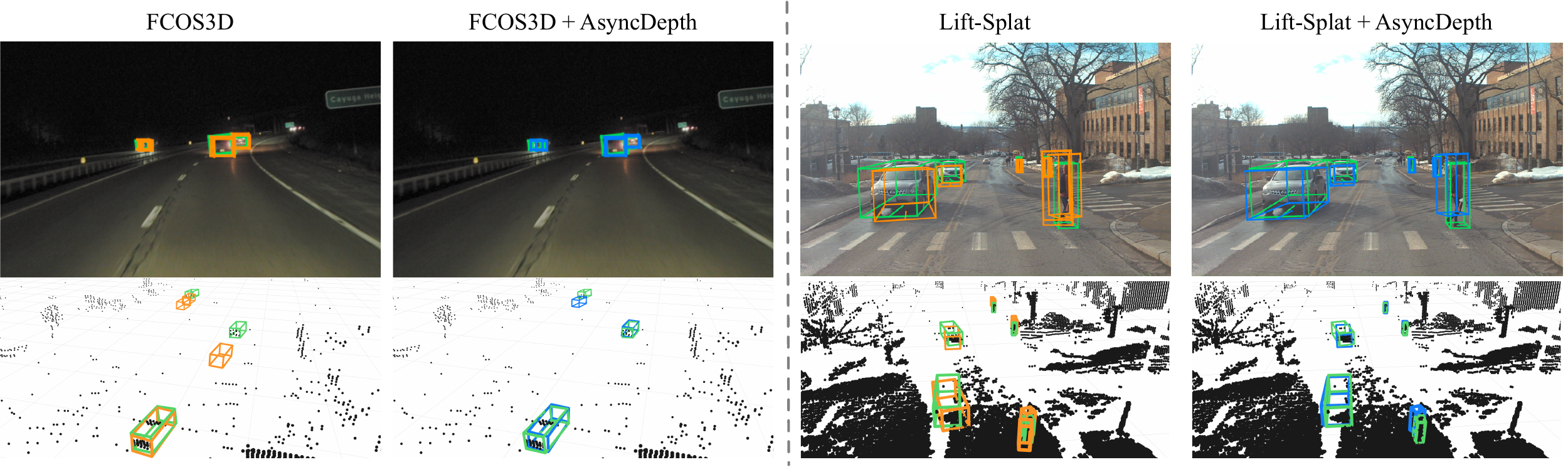}
    \caption{\textbf{Qualitative visualizations of \methodshort.} We visualize 3D detection in the monocular image on the \ith dataset. Ground truth boxes are shown in \textcolor{iosgreen}{green}, baseline model predictions are shown in \textcolor{iosorange}{orange}, and \methodshort predictions are shown in \textcolor{iosblue}{blue}. We also include --for visualization purposes only-- the detections in 3D overlaid with the LiDAR point cloud (note: these are \textbf{not} given as the model inputs). Observe that, in the image, the 3D detections are ambiguous and the depth is incorrect in the baseline models. \methodshort is able to correct the depth to produce more accurate 3D detections.}
    \label{fig:qualitiative}
    \vspace{-10pt}
\end{figure*}

We visualize the detection results on \ith in \autoref{fig:qualitiative}. Though the base detector is able to roughly estimate the location of the object, the \methodshort variant is able to accurately localize the object, suggesting that \methodshort is able to improve the base detector. 

\section{Analysis: Effects of localization errors.}
\label{sec: localization_errors}
\input{tables/localization_errors.tex}
\methodshort requires accurate localization to query the past traversal LiDAR scans. Thus it is crucial to analyze its robustness against errors in localization during inference. To simulate localization error, we randomly perturb the localization of the ego vehicle by a vector $\epsilon \mathbf{r}$ where $\epsilon$ is a scalar sampled from $\mathcal{N}(0, \sigma_t^2)$ and $\mathbf{r}$ is unit vector randomly sampled from the 3D space. In addition, we 
also simulate random heading error by injecting 
random noises sampled from $\mathcal{N}(0, \sigma_r^2)$ to the yaw angle of the vehicle. We report the result in \autoref{tab: localization errors}.
We observe that the translation error has to stay within noise level $\sigma_t = 0.2$m for \methodshort to maintain improvements over the baselines. 
Note that this is not difficult to achieve with current localization technology.
\methodshort is quite robust to noise in yaw angles. Note that our choice of yaw noise level is much harsher than typically expected noise since with current technology the error can be as small as 0.08$^{\circ }$ \cite{compact_dual-antenna}.

\section{Analysis: Better Depth Estimation from \methodshort.} 
\begin{figure}[!t]
    \centering
    \includegraphics[width=\linewidth]{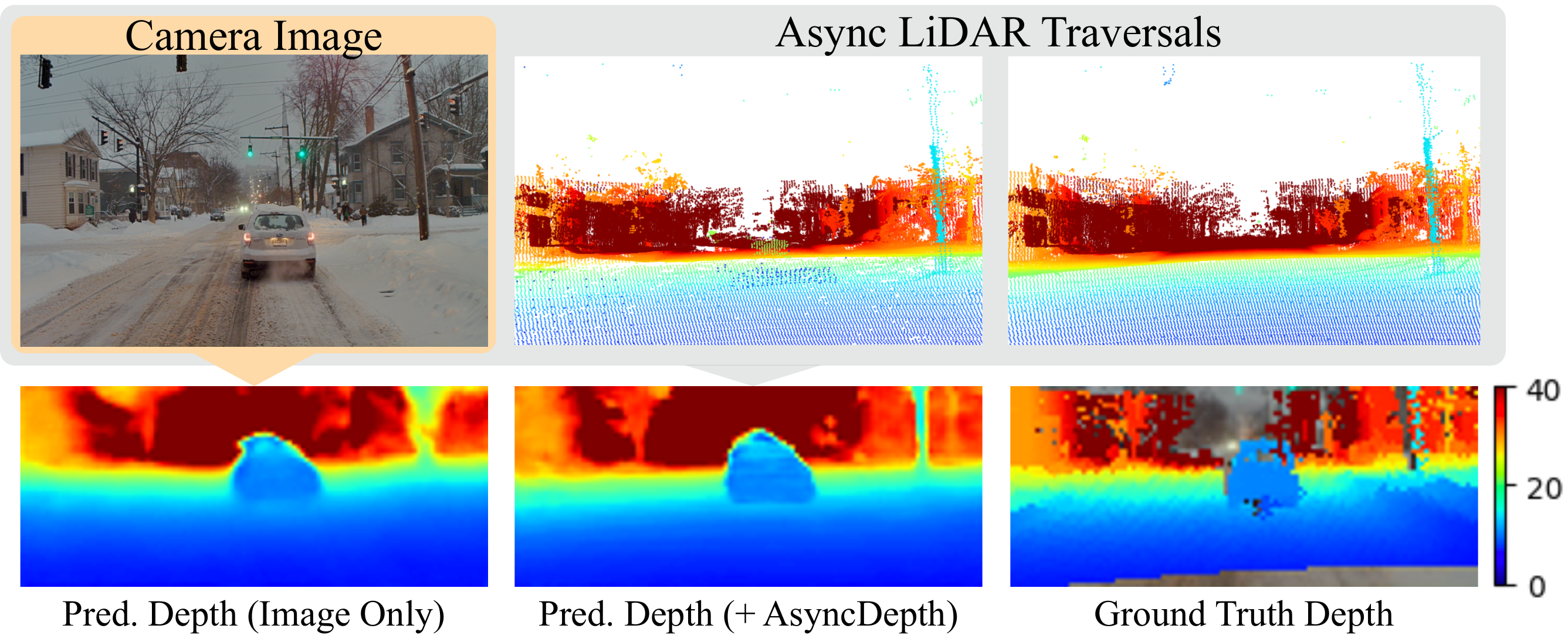}
    \caption{\textbf{Visualizing the depth predictions on the Lift-Splat model.} We show the current image (top left) and two of the corresponding asynchronous depth maps (top right). 
    On the bottom, we show the predicted depth inside the Lift-Splat model with and without \methodshort.
    We show the ground-truth synchronous depth map on the bottom right as a reference.
    The colorbar indicates the corresponding depth in meters.
    Best viewed in color.
    }
    \label{fig:qualdepth}
    \vspace{-15pt}
\end{figure}

We posit that the gain in detection performance could be partly attributed to improved depth estimation with extra \methodshort information. Qualitatively, we visualize the depth map produced internally by Lift-Splat in \autoref{fig:qualdepth}. We observe that with \methodshort, the model is able to produce much better depth estimates for the whole scene, especially for static objects (compare the shape of the phone pole and tree in the first two depth maps from the left) and the ground plane. Quantitatively, we compute the L1 depth error for the depth estimation from the model with and without \methodshort, and it yields 1.52 and 2.16 respectively, aligning with our observation.

\begin{figure}[!t]
    \centering
    \includegraphics[width=.85\linewidth]{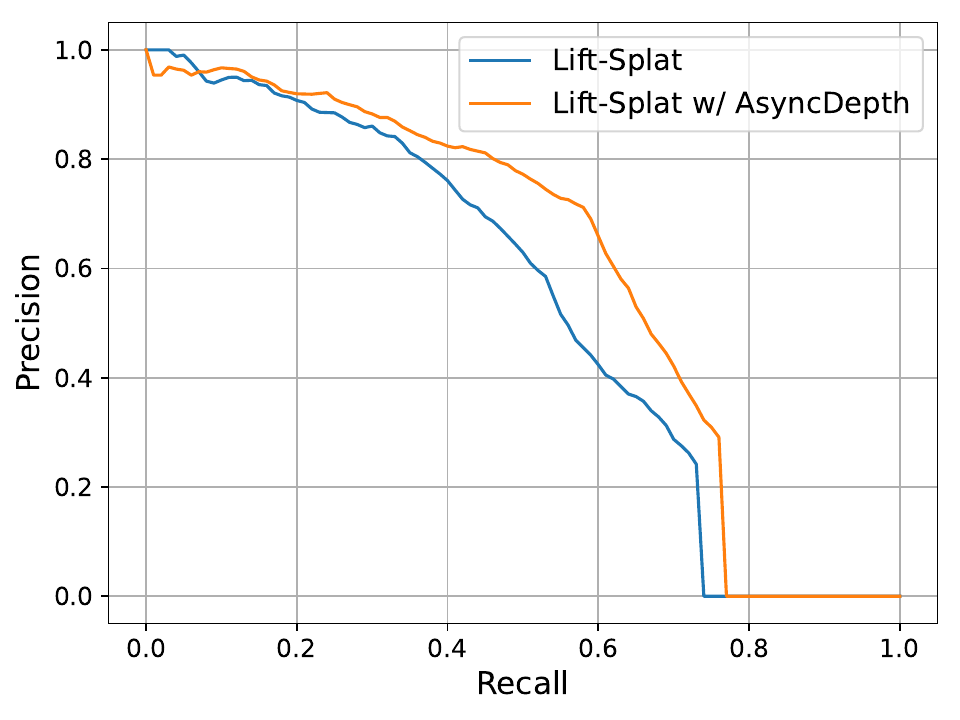}
    \vspace{-5pt}
    \caption{\textbf{Precision-recall curve comparing the baseline model and \methodshort variant for car detection on \ith (distance threshold = 1m).} With \methodshort, the detector makes more precise predictions while maintaining similar recall. }
    \label{fig:pr_curve}
    \vspace{-5pt}
\end{figure}
\section{Qualitative Analysis: Precision-Recall Curve}
We visualize the precision-recall curve on \ith for car detection (the class where we observe the most improvement) in \autoref{fig:pr_curve}. We observe that with \methodshort, the detector can produce more accurate predictions of the car.

\section{Additional Evaluation Results}
We report the performance under different evaluation metrics on \lyft in \autoref{tab:lyft_DS}, \autoref{tab:lyft_ATE}, \autoref{tab:lyft_ASE}, \autoref{tab:lyft_AOE}. Results on \ith can be found in \autoref{tab:ithaca_DS}, \autoref{tab:ithaca_ATE}, \autoref{tab:ithaca_ASE}, \autoref{tab:ithaca_AOE}. For the detailed definition of Average Translation Error (ATE), Average Scale Error (ASE), Average Orientation Error (AOE), please refer to \cite{caesar2020nuscenes}. The definition of Detection Score (DS) can be found in the main text. 
\input{tables/lyft_supp.tex}

\input{tables/ithaca_supp.tex}

%% file: tables/localization_errors.tex
\begin{table}[!t]
    \small
    \centering
    \resizebox{.47\textwidth}{!}{%
    \tabcolsep 4pt
    \begin{tabular}{ccccccccc}
        \toprule 
        \multicolumn{2}{c}{Loc. Error} & \multicolumn{1}{c}{\multirow{2}{*}{mAP}} & \multicolumn{3}{c}{Car} &  \multicolumn{3}{c}{Pedestrian} \\
        \cmidrule(lr){4-6} \cmidrule(lr){7-9}
        Trans. & Rot. & & 0-30 & 30-50 & 0-50 & 0-30 & 30-50 & 0-50  \\
        \midrule
        \multicolumn{2}{c}{baseline} & 39.4 & 66.6 & 30.2 & 52.8 & 37.1 & 13.7 & 26.0 \\
        \midrule
        0.0m & 0.0$^\circ$ & 42.9 & 70.2 & 38.2 & 58.3 & 37.6 & 16.6 & 27.5\\
        0.1m & 0.0$^\circ$ & 41.8 & 68.8 & 36.5 & 56.8 & 36.8 & 16.5 & 26.7\\
        0.2m & 0.0$^\circ$ & 39.4 & 66.0 & 33.3 & 53.8 & 34.1 & 15.6 & 25.0\\
        0.0m & 0.1$^\circ$ & 42.8 & 69.9 & 38.1 & 58.1 & 37.4 & 16.5 & 27.6\\
        0.0m & 1.0$^\circ$ & 42.5 & 70.0 & 36.9 & 57.5 & 37.2 & 16.8 & 27.6\\
        0.1m & 1.0$^\circ$ & 40.9 & 68.0 & 34.8 & 55.5 & 36.0 & 15.6 & 26.3\\
        \bottomrule
    \end{tabular}
    }
    \caption{\textbf{Mean Average Precision of using \methodshort with various simulated localization errors.} The underlining model is trained without localization errors, and we simulate various localization errors during testing.}
    \label{tab: localization errors}
    \vspace{-15pt}
\end{table}

%% file: tables/lyft_supp.tex
\begin{table}[!t]
    \centering
    \tabcolsep 4pt
    \resizebox{.3\textwidth}{!}{%
    \begin{tabular}{@{}ccc@{}}
        \toprule 
        Method  & FCOS3D~\cite{FCOS3D2021} & Lift-Splat~\cite{LSS2020} \\
        \cmidrule{1-3} 
        Baseline & 0.267& 0.290\\
        + \methodshort  & 0.272 & 0.327\\
        
        \cmidrule{2-3}
        $\Delta$ DS & \textcolor{TableGreen}{0.005} & \textcolor{TableGreen}{0.037}\\

        \bottomrule
    \end{tabular}
    }
    \caption{\textbf{Detection Score (DS) of two types of detectors across different ranges and object class types on the \lyft dataset (higher is better)}. 
    }
    \label{tab:lyft_DS}
    \vspace{-5pt}
\end{table}

\begin{table}[!t]
    \centering
    \tabcolsep 4pt
    \resizebox{.3\textwidth}{!}{%
    \begin{tabular}{@{}ccc@{}}
        \toprule 
        Method  & FCOS3D~\cite{FCOS3D2021} & Lift-Splat~\cite{LSS2020} \\
        \cmidrule{1-3} 
        Baseline & 0.339& 0.393\\
        + \methodshort  & 0.377 & 0.423\\
        
        \cmidrule{2-3}
        $\Delta$ DS & \textcolor{TableGreen}{0.038} & \textcolor{TableGreen}{0.030}\\

        \bottomrule
    \end{tabular}
    }
    \caption{\textbf{Detection Score (DS) of two types of detectors across different ranges and object class types on the \ith dataset (higher is better)}. 
    }
    \label{tab:ithaca_DS}
    \vspace{-10pt}
\end{table}

\begin{table*}[!t]
    \tabcolsep 4pt
    \centering
    \resizebox{\textwidth}{!}{%
    \begin{tabular}{@{}ccccccccccccccccc@{}}
        \toprule 
        \multicolumn{1}{c}{\multirow{2}{*}{Method}} & \multicolumn{1}{c}{\multirow{2}{*}{ATE}} &  \multicolumn{3}{c}{Car} & \multicolumn{3}{c}{Truck} & \multicolumn{3}{c}{Bus} & \multicolumn{3}{c}{Bicycle} &  \multicolumn{3}{c}{Pedestrian} \\
        \cmidrule(lr){3-5} \cmidrule(lr){6-8} \cmidrule(lr){9-11} \cmidrule(l){12-14} \cmidrule(l){15-17}
        & & 0-30 & 30-50 & 0-50 & 0-30 & 30-50 & 0-50 & 0-30 & 30-50 & 0-50 & 0-30 & 30-50 & 0-50 & 0-30 & 30-50 & 0-50 \\
        \midrule
        FCOS3D~\cite{FCOS3D2021} & 0.97 & 0.75 &0.91&	0.81&	1.10&	1.03&	1.07&	1.15&	1.06&	1.11&	0.97&	0.96&	0.97&	0.87&	0.91& 0.89 \\
        + \methodshort & 0.99 & 0.75 & 0.91 & 0.81 & 1.05 & 1.26 & 1.16 & 1.13 & 1.16 & 1.17 & 0.90 & 0.96 & 0.92 & 0.88 & 0.94 & 0.91 \\
        \cmidrule{2-17}
        $\Delta$ ATE & \textcolor{red}{0.02} & 0.00 & 0.00 & 0.00 &	\textcolor{TableGreen}{-0.05}&	\textcolor{red}{0.23} &	\textcolor{red}{0.09} &	\textcolor{TableGreen}{-0.02}&	\textcolor{red}{0.10} &	\textcolor{red}{0.06} &	\textcolor{TableGreen}{-0.07}&	 0.00 &	\textcolor{TableGreen}{-0.05}&	\textcolor{red}{0.01} &	\textcolor{red}{0.03}&	\textcolor{red}{0.02} \\
        \midrule
        Lift-Splat~\cite{LSS2020} & 0.83 & 0.43&	0.85&	0.55&	1.05&	1.06&	1.07&	1.18&	1.10&	1.13&	0.60&	1.01&	0.71&	0.64&	0.84&	0.72 \\
        + \methodshort & 0.78 & 0.41&	0.86&	0.54&	1.08&	0.96&	1.03&	1.00&	0.94&	0.98&	0.55&	1.02&	0.67 &	0.61&	0.75&	0.68 \\
         \cmidrule{2-17}
        $\Delta$ ATE & \textcolor{TableGreen}{-0.05}& \textcolor{TableGreen}{-0.02}&	\textcolor{red}{0.01}&	\textcolor{TableGreen}{-0.01}&	\textcolor{red}{0.03} &	\textcolor{TableGreen}{-0.10}&	\textcolor{TableGreen}{-0.04}&	\textcolor{TableGreen}{-0.18} &	\textcolor{TableGreen}{-0.16}&	\textcolor{TableGreen}{-0.15}&	\textcolor{TableGreen}{-0.05}&	\textcolor{red}{0.01} &	\textcolor{TableGreen}{-0.04}&	\textcolor{TableGreen}{-0.03}&	\textcolor{TableGreen}{-0.09}&	\textcolor{TableGreen}{-0.04}\\

        \bottomrule
        \end{tabular}
    }
    \caption{\textbf{Average translation error (ATE) of two types of detectors across different ranges and object class types on the \lyft dataset for all the true positive detections within 2m (lower is better).} 
    We evaluate two types of monocular 3D object detection models (FCOS3D~\cite{FCOS3D2021} and Lift-Splat~\cite{LSS2020,reading2021categorical,liu2022bevfusion,huang2021bevdet}) with and without \methodshort. 
    We show the ATE metric and its breakdown across different ranges (in meters) and class objects. 
    ``$\Delta$ ATE'' indicates the difference between the detectors without and with \methodshort.
    }
    \label{tab:lyft_ATE}
\end{table*}

\begin{table*}[!t]
    \tabcolsep 4pt
    \centering
    \resizebox{\textwidth}{!}{%
    \begin{tabular}{@{}ccccccccccccccccc@{}}
        \toprule 
        \multicolumn{1}{c}{\multirow{2}{*}{Method}} & \multicolumn{1}{c}{\multirow{2}{*}{ASE}} &  \multicolumn{3}{c}{Car} & \multicolumn{3}{c}{Truck} & \multicolumn{3}{c}{Bus} & \multicolumn{3}{c}{Bicycle} &  \multicolumn{3}{c}{Pedestrian} \\
        \cmidrule(lr){3-5} \cmidrule(lr){6-8} \cmidrule(lr){9-11} \cmidrule(l){12-14} \cmidrule(l){15-17}
        & & 0-30 & 30-50 & 0-50 & 0-30 & 30-50 & 0-50 & 0-30 & 30-50 & 0-50 & 0-30 & 30-50 & 0-50 & 0-30 & 30-50 & 0-50 \\
        \midrule
        FCOS3D~\cite{FCOS3D2021} & 0.23 & 0.13&	0.14&	0.13&	0.25&	0.24&	0.24&	0.21&	0.24&	0.22&	0.26&	0.33&	0.27&	0.28&	0.25&	0.28\\
        + \methodshort & 0.24 & 0.12 & 0.14 & 0.13 & 0.28 & 0.25 & 0.26 & 0.21 & 0.27 & 0.25 & 0.25 & 0.33 & 0.26 & 0.29 & 0.35 & 0.30 \\
        \cmidrule{2-17}
        $\Delta$ ASE & \textcolor{red}{0.01} & \textcolor{TableGreen}{-0.01} & 0.00 &	0.00	& \textcolor{red}{0.03}& \textcolor{red}{0.01} & 	\textcolor{red}{0.02} &	0.00 & \textcolor{red}{0.03} & \textcolor{red}{0.03} &	\textcolor{TableGreen}{-0.01}&	 0.00 &	\textcolor{TableGreen}{-0.01}& \textcolor{red}{0.01} &	\textcolor{red}{0.10}& \textcolor{red}{0.02}  \\
        \midrule
        Lift-Splat~\cite{LSS2020} & 0.24 & 0.15 &	0.17&	0.16&	0.30&	0.28&	0.29&	0.25&	0.28&	0.27&	0.22&	0.30&	0.24&	0.26&	0.22&	0.25 \\
        + \methodshort & 0.24 & 0.15&	0.17&	0.15&	0.29&	0.28&	0.29&	0.23&	0.27&	0.25&	0.23&	0.27&	0.24&	0.26&	0.21&	0.25 \\
         \cmidrule{2-17}
        $\Delta$ ASE & 0.00 & 0.00 & 0.00 &	\textcolor{TableGreen}{-0.01}&	\textcolor{TableGreen}{-0.01}& 0.00 & 0.00 &	\textcolor{TableGreen}{-0.02}&	\textcolor{TableGreen}{-0.01}&	\textcolor{TableGreen}{-0.02}&	\textcolor{red}{0.01}&	\textcolor{TableGreen}{-0.03}& 0.00 & 0.00 &	\textcolor{TableGreen}{-0.01}& 0.00  \\
        \bottomrule
    \end{tabular}
    }
    \caption{\textbf{Average scale error (ASE) of two types of detectors across different ranges and object class types on the \lyft dataset for all the true positive detections within 2m (lower is better).} 
    We evaluate two types of monocular 3D object detection models (FCOS3D~\cite{FCOS3D2021} and Lift-Splat~\cite{LSS2020,reading2021categorical,liu2022bevfusion,huang2021bevdet}) with and without \methodshort. 
    We show the ASE metric and its breakdown across different ranges (in meters) and class objects. 
    ``$\Delta$ ASE'' indicates the difference between the detectors without and with \methodshort.
    }
    \label{tab:lyft_ASE}
\end{table*}

\begin{table*}[!t]
    \tabcolsep 4pt
    \centering
    \resizebox{\textwidth}{!}{%
    \begin{tabular}{@{}ccccccccccccccccc@{}}
        \toprule 
        \multicolumn{1}{c}{\multirow{2}{*}{Method}} & \multicolumn{1}{c}{\multirow{2}{*}{AOE}} &  \multicolumn{3}{c}{Car} & \multicolumn{3}{c}{Truck} & \multicolumn{3}{c}{Bus} & \multicolumn{3}{c}{Bicycle} &  \multicolumn{3}{c}{Pedestrian} \\
        \cmidrule(lr){3-5} \cmidrule(lr){6-8} \cmidrule(lr){9-11} \cmidrule(l){12-14} \cmidrule(l){15-17}
        & & 0-30 & 30-50 & 0-50 & 0-30 & 30-50 & 0-50 & 0-30 & 30-50 & 0-50 & 0-30 & 30-50 & 0-50 & 0-30 & 30-50 & 0-50 \\
        \midrule
        FCOS3D~\cite{FCOS3D2021} & 0.63 & 0.06 &	0.07&	0.06&	0.24&	0.12&	0.18&	0.27&	0.51&	0.40&	1.03&	1.76&	1.24&	1.14&	1.74&	1.28 \\
        + \methodshort & 0.61 & 0.08 & 0.08 & 0.07 & 0.18 & 0.12 & 0.16 & 0.40 & 0.35 & 0.34 & 1.15 & 1.21 & 1.16 & 1.24 & 1.54 & 1.34 \\
        \cmidrule{2-17}
        $\Delta$ AOE & \textcolor{TableGreen}{-0.02} &\textcolor{red}{0.02} &	\textcolor{red}{0.01}&	\textcolor{red}{0.01}&	\textcolor{TableGreen}{-0.06}& 0.00 &	\textcolor{TableGreen}{-0.02}& \textcolor{red}{0.13} &	\textcolor{TableGreen}{-0.16}&	\textcolor{TableGreen}{-0.06}&	\textcolor{red}{0.12}&	\textcolor{TableGreen}{-0.55}&	\textcolor{TableGreen}{-0.08}&	\textcolor{red}{0.10}&	\textcolor{TableGreen}{-0.20}&	\textcolor{red}{0.06} \\
        \midrule
        Lift-Splat~\cite{LSS2020} & 0.85 & 0.15 &	0.30&	0.20&	0.63&	0.45&	0.55&	0.19&	0.36&	0.28&	1.50&	1.31&	1.45&	1.73&	2.18&	1.78 \\
        + \methodshort & 0.78 & 0.13&	0.30&	0.18&	0.48&	0.43&	0.46&	0.10&	0.18&	0.14&	1.63&	1.45&	1.58&	1.54&	1.60&	1.55 \\
         \cmidrule{2-17}
        $\Delta$ AOE & \textcolor{TableGreen}{-0.07}& \textcolor{TableGreen}{-0.02}& 0.00 &	\textcolor{TableGreen}{-0.02}&	\textcolor{TableGreen}{-0.15}&	\textcolor{TableGreen}{-0.02}&	\textcolor{TableGreen}{-0.09}&	\textcolor{TableGreen}{-0.09}&	\textcolor{TableGreen}{-0.18}&	\textcolor{TableGreen}{-0.14}&	\textcolor{red}{0.13}&	\textcolor{red}{0.14}&	\textcolor{red}{0.13}&	\textcolor{TableGreen}{-0.19}&	\textcolor{TableGreen}{-0.58}&	\textcolor{TableGreen}{-0.23} \\
        \bottomrule
    \end{tabular}
    }
    \caption{\textbf{Average orientation error (AOE) of two types of detectors across different ranges and object class types on the \lyft dataset for all the true positive detections within 2m (lower is better).} 
    We evaluate two types of monocular 3D object detection models (FCOS3D~\cite{FCOS3D2021} and Lift-Splat~\cite{LSS2020,reading2021categorical,liu2022bevfusion,huang2021bevdet}) with and without \methodshort. 
    We show the AOE metric and its breakdown across different ranges (in meters) and class objects. 
    ``$\Delta$ AOE'' indicates the difference between the detectors without and with \methodshort.
    }
    \label{tab:lyft_AOE}
\end{table*}

%% file: tables/ithaca_supp.tex
\begin{table}[!t]
    \centering
    \tabcolsep 4pt
    \resizebox{.47\textwidth}{!}{%
    \begin{tabular}{@{}cccccccc@{}}
        \toprule 
        \multicolumn{1}{c}{\multirow{2}{*}{Method}} & \multicolumn{1}{c}{\multirow{2}{*}{mAP}} & \multicolumn{3}{c}{Car} &  \multicolumn{3}{c}{Pedestrian} \\
        \cmidrule(lr){3-5} \cmidrule(lr){6-8}
        & & 0-30 & 30-50 & 0-50 & 0-30 & 30-50 & 0-50  \\
        \midrule
        FCOS3D~\cite{FCOS3D2021} & 0.89 & 0.63 & 0.93 & 0.74 & 0.96 & 1.08 & 1.04  \\
        + \methodshort & 0.82 & 0.53 & 0.86 & 0.66 & 0.85 & 1.05 & 0.98\\
        \cmidrule{2-8}
        $\Delta$ ATE & \textcolor{TableGreen}{-0.07} & \textcolor{TableGreen}{-0.10} & \textcolor{TableGreen}{-0.07} & \textcolor{TableGreen}{-0.09} & \textcolor{TableGreen}{-0.11} & \textcolor{TableGreen}{-0.04} & \textcolor{TableGreen}{-0.06}  \\
        \midrule
        Lift-Splat~\cite{LSS2020} & 0.77  & 0.46 & 0.89 & 0.58 & 0.85 & 1.04 & 0.96 \\
        + \methodshort & 0.70  & 0.41 & 0.79 & 0.52 & 0.75 & 0.99 & 0.87 \\
        \cmidrule{2-8}
        $\Delta$ ATE & \textcolor{TableGreen}{-0.07}  & \textcolor{TableGreen}{-0.05} & \textcolor{TableGreen}{-0.10} & \textcolor{TableGreen}{-0.06} & \textcolor{TableGreen}{-0.10} & \textcolor{TableGreen}{-0.06} & \textcolor{TableGreen}{-0.09} \\
        \bottomrule
    \end{tabular}
    }
    \caption{\textbf{Average translation error (ATE) of two types of detectors across different ranges and object class types on the \ith dataset for all the true positive detections within 2m (lower is better).} Please refer to \autoref{tab:lyft_map} for naming.
    }
    \label{tab:ithaca_ATE}
\end{table}

\begin{table}[!t]
    \centering
    \tabcolsep 4pt
    \resizebox{.47\textwidth}{!}{%
    \begin{tabular}{@{}cccccccc@{}}
        \toprule 
        \multicolumn{1}{c}{\multirow{2}{*}{Method}} & \multicolumn{1}{c}{\multirow{2}{*}{ASE}} & \multicolumn{3}{c}{Car} &  \multicolumn{3}{c}{Pedestrian} \\
        \cmidrule(lr){3-5} \cmidrule(lr){6-8}
        & & 0-30 & 30-50 & 0-50 & 0-30 & 30-50 & 0-50  \\
        \midrule
        FCOS3D~\cite{FCOS3D2021} & 0.18 & 0.10 & 0.07 & 0.09 & 0.26 & 0.29 & 0.28  \\
        + \methodshort & 0.19 & 0.10 & 0.07 & 0.09 & 0.27 & 0.30 & 0.28 \\
        \cmidrule{2-8}
        $\Delta$ ASE & \textcolor{red}{0.01} & 0.00 & 0.00 & 0.00 & \textcolor{red}{0.01} & 0.00 & \textcolor{red}{0.01}  \\
        \midrule
        Lift-Splat~\cite{LSS2020} &  0.19 & 0.12 & 0.08 & 0.11 & 0.25 & 0.28 & 0.27 \\
        + \methodshort & 0.19 & 0.12 & 0.09 & 0.11 & 0.26 & 0.28 & 0.27 \\
        \cmidrule{2-8}
        $\Delta$ ASE & 0.00 & 0.00 & \textcolor{red}{0.01} & 0.00 & 0.00 & 0.00 & 0.00  \\
        \bottomrule
    \end{tabular}
    }
    \caption{\textbf{Average scale error (ASE) of two types of detectors across different ranges and object class types on the \ith dataset for all the true positive detections within 2m (lower is better).} 
    Please refer to \autoref{tab:lyft_map} for naming.
    }
    \label{tab:ithaca_ASE}
\end{table}

\newpage
\begin{table}[!t]
    \centering
    \tabcolsep 4pt
    \resizebox{.47\textwidth}{!}{%
    \begin{tabular}{@{}cccccccc@{}}
        \toprule 
        \multicolumn{1}{c}{\multirow{2}{*}{Method}} & \multicolumn{1}{c}{\multirow{2}{*}{AOE}} & \multicolumn{3}{c}{Car} &  \multicolumn{3}{c}{Pedestrian} \\
        \cmidrule(lr){3-5} \cmidrule(lr){6-8}
        & & 0-30 & 30-50 & 0-50 & 0-30 & 30-50 & 0-50  \\
        \midrule
        FCOS3D~\cite{FCOS3D2021} &0.64 & 0.10 & 0.10 & 0.10 & 1.07 & 1.31 & 1.18  \\
        + \methodshort & 0.61 & 0.10 & 0.09 & 0.10 & 1.01 & 1.23 & 1.12 \\
        \cmidrule{2-8}
        $\Delta$ AOE & \textcolor{TableGreen}{-0.03} & 0.00 & \textcolor{TableGreen}{-0.02} & 0.00 & \textcolor{TableGreen}{-0.06} & \textcolor{TableGreen}{-0.07} & \textcolor{TableGreen}{-0.06}  \\
        \midrule
        Lift-Splat~\cite{LSS2020} & 0.87 & 0.20 & 0.34 & 0.26 & 1.50 & 1.41 & 1.47 \\
        + \methodshort & 0.85 & 0.24 & 0.34 & 0.29 & 1.42 & 1.45 & 1.42 \\
        \cmidrule{2-8}
        $\Delta$ AOE & \textcolor{TableGreen}{-0.02}  & \textcolor{red}{0.04} & 0.00 & \textcolor{red}{0.03} & \textcolor{TableGreen}{-0.08} & \textcolor{red}{0.04} & \textcolor{TableGreen}{-0.05} \\
        \bottomrule
    \end{tabular}
    }
    \caption{\textbf{Average orientation error (AOE) of two types of detectors across different ranges and object class types on the \ith dataset for all the true positive detections within 2m (lower is better).} 
    Please refer to \autoref{tab:lyft_map} for naming.
    }
    \label{tab:ithaca_AOE}
\end{table}